\documentclass[conference]{IEEEtran}
\IEEEoverridecommandlockouts
\usepackage{cite}
\usepackage{amsmath,amssymb,amsfonts}
\usepackage{algorithmic}
\usepackage{graphicx}
\usepackage{textcomp}
\usepackage{xcolor}
\usepackage{pifont}
\usepackage{caption}
\usepackage{comment}

\usepackage{fancyhdr}
\def\BibTeX{{\rm B\kern-.05em{\sc i\kern-.025em b}\kern-.08em
    T\kern-.1667em\lower.7ex\hbox{E}\kern-.125emX}}
\begin{document}

\title{ITHACA. A TOOL FOR INTEGRATING FUZZY LOGIC IN UNITY}

\author{\IEEEauthorblockN{1\textsuperscript{st} Alfonso Tejedor Moreno}
\IEEEauthorblockA{\textit{Department of Computer Science} \\
\textit{University of Almeria}\\
Almeria 04120, Spain \\
alfitejedor@gmail.com}
\and
\IEEEauthorblockN{2\textsuperscript{nd} Jose A. Piedra-Fernandez}
\IEEEauthorblockA{\textit{Department of Computer Science} \\
\textit{University of Almeria}\\
Almeria 04120, Spain  \\
jpiedra@ual.es}
\and
\IEEEauthorblockN{3\textsuperscript{rd} Juan Jesus Ojeda-Castelo}
\IEEEauthorblockA{\textit{Department of Computer Science} \\
\textit{University of Almeria}\\
Almeria 04120, Spain \\
juanje.ojeda@ual.es}
\and
\IEEEauthorblockN{4\textsuperscript{th} Luis Iribarne \IEEEmembership{Member,~IEEE}}
\IEEEauthorblockA{\textit{Department of Computer Science} \\
\textit{University of Almeria}\\
Almeria 04120, Spain \\
luis.iribarne@ual.es}
}

\maketitle

\begin{abstract}
Ithaca is a Fuzzy Logic (FL) plugin for developing  artificial intelligence systems within the Unity game engine. Its goal is to provide an intuitive and natural way to build advanced artificial intelligence systems, making the implementation of such a system faster and more affordable. The software is made up by a C\# framework and an Application Programming Interface (API) for writing inference systems, as well as a set of tools for graphic development and debugging. Additionally, a Fuzzy Control Language (FCL) parser is provided in order to import systems previously defined using this standard.
\end{abstract}


\section{Introduction}
\IEEEPARstart{G}{raphics} is a relevant factor in terms of video games and that fact is appreciated in the latest development of video games and consoles. Each new generation of consoles empower their resources such as a graphic card with a lot of calculation, a processor with many cores and so forth. Therefore, the video games are developed with very realistic graphics and the users feel that they are part of a real environment more than a virtual world. However, users are more demanding and currently artificial intelligence (AI) plays a significance role in this field since a video game is not really popular if its graphics are spectacular but its AI is very low. In fact, the quality of AI is an important feature in order to increase the immersion and enjoyment of the players \cite{nareyek2004ai}.

 The AI has been a magnificent contribution to the video games sector since it has mastered games as Chess \cite{newborn2012kasparov} or Go \cite{silver2016mastering}. Or even the algorithm developed by DeepMind which is able to cooperate with human players or other machine players \cite{jaderberg2019human}. These agents have been tested in the game Quake III.  According to Pirovano \cite{Pirovano12}, the two most common AI techniques found in games are Finite State Machines (FSM) and Decision Trees (DT), far from the performance of more complex techniques like Artificial Neural Networks (ANN) and Genetic Algorithms (GA). Therefore our main goal is making AI achievable for small and medium sized teams, providing a tool for implementing a technique halfway between the easy to use of the FSM and the power and flexibility of the ANN. And right from the most used game engine, Unity \footnote{Unity - https://unity.com/} (Fast Facts - Unity) The AI in games is not only for the NPC or that the player can make more decisions in the game if not that AI is being used to remaster video games and improve the quality of these products in hours. 
 
 On the other hand, most game developers work on huge constraints of time and budget, thus the AI implemented specifically in games by indie developers is naive at best, if there is any at all. That is the reason that we have developed a plugin with fuzzy logic in order to help the developers and so forth to apply fuzzy logic in their projects in a easy way and make the most of it. Moreover, this plugin was developed for Unity because this game engine is operated from students or hobbyists to professionals and game studios. Unity allows the developers to create video games in multiple platforms, more exactly 30 different platforms, such as mobile, PC or console. As a result, this tool is very useful for people who wants to learn how to develop games, people that they only want to create a video game as a hobby or even in a more professional level entrepreneurs who wants to set up their entertainment company or create a professional game. Furthermore, this plugin has been designed with the fundamental functions of fuzzy logic and is very easy to use because it is aimed at beginners without previous knowledge of the subject. And this is possible because Unity is adapted to many needs being its learning process very easy at the beginning although it gets more difficult if you want to do a complex game. These features make that this game engine is used by a wider population than other game engines as Unreal Engine \footnote{Unreal Engine - https://www.unrealengine.com/en-US/} In fact,  John Riccitiello (Unity CEO) states that the half of all the games has been developed on Unity \footnote{Unity games developed - https://techcrunch.com/2018/09/05/unity-ceo-says-half-of-all-games-are-built-on-unity/}.

The objectives of the project are:

\begin{itemize}
\item Create an easy to use tool that makes the Artificial Intelligence and Fuzzy Logic available to all game developers.
\item The tool has to be simple and flexible in order to be used in any kind of projects and games.
\item The user must be able to work with the tool using either a Graphical User Interface (GUI) or an API.
\item Be compliant with the FCL standard as defined in IEC 61131-7 \cite{iec}
\end{itemize}

The remainder of the paper is organized as follows: Section \ref{stateArt} presents
some background on Fuzzy Logic and Artificial Intelligence. Section \ref{proposal} describes the architecture and its Fuzzy Logic foundations. Section \ref{results} shows the results obtained by building a few small games.
Section \ref{conclusions} summarizes the conclusions and discusses the future work.

\section{Background} \label{stateArt}

In 1940 Edward Uhler Condon presented a Math game at the exhibition of Westinghouse. This human vs machine game was played by thousands of people of which only 10\% were able to beat the machine. However, it was not until the 1970s that video games began to commercialize on a large scale, as in the case of the game developed by the Atari company: Pong \cite{perry1983microprocessors}. At that time, Neural Networks were not included in the AI system of the video game, although simpler techniques such as finite state machines were used. The next section will briefly describe the influence of AI in the world of video games.

\subsection{Artificial Intelligence in Games}

In 1959 Arthur Lee Samuel introduced Artificial Intelligence into the world of video games, more specifically teaching a computer how to play the game of checkers \cite{samuel1959some}. Thereafter, other games were developed that included artificial intelligence, highlighting the video game Space Invaders in the 70s and Pac-Man in the 80s. In Space Invaders, complexity was added to the game by introducing hash functions that "learned" from the player's actions. The heart of the Pac-Man game is the behavior of the ghosts that are the player's enemies, where each one has a different behavior \cite{lammers1989programmers} and it is controlled by a state machine \cite{millington2019ai}. In the 90s video games were booming due to the appearance of various consoles such as Gameboy, SNES or PS One. This expansion also produced the inclusion of more complex AI systems, with special mention to the Half-life video game in which it existed a cooperation between the enemies to kill the player \cite{birdwell1999valve}. In the 21st century AI has advanced considerably and its algorithms are capable of defeating a human player in games like Starcraft II \cite{vcerticky2018starcraft, vinyals2017starcraft}.

All these developments have in common, regardless of the AI that they integrated, the improvement of the user experience and for this objective AI is an outstanding candidate. AI can be used in different aspects of a video game as pathfinding, movement, making decision or learning \cite{millington2019ai}, but probably the one that stands out the most is the one that controls the behavior of NPCs. The reason is because if the behavior of these elements is realistic and resembles the real world, the users will feel that he is not playing in an artificial and predefined world but in a world more similar to their own. In reference to this last idea, it is important to remember that the aim of the AI is not to beat the player always, since the fun and motivation would be lost, but rather to create an experience as realistic as possible.

There are even studies that have analyzed the feasibility of Deep reinforcement learning for serious games \cite{dobrovsky2017deep}

In addition to Deep reinforcement learning, genetic algorithms have brought improvements in this area. In \cite{martinez2016creating} AI characters for fighting games can be generated with the help of genetic algorithms and through a low-cost process. Moreover, it has also been used in a soccer simulator (Robocup Soccer) with the aim of improving decision making in the simulation of soccer teams \cite{tavafi2017hybrid}. The result of the implementation of this enhancement was that two of the three teams that made up this system were top teams in the RoboCup competition. In \cite{marino2019learning} sets out to apply genetic algorithms in real-time strategy games to learn new and effective strategies for this type of game. The problem is that these types of actions are usually abstract for the player and for this reason in this work they focus on learning strategies that can be easily interpreted by people. Furthermore, this type of algorithms have also been used in games such as Backgammon \cite{azaria2005gp} or Chess \cite{hauptman2005gp}.

In the previous paragraphs, a series of applications have been shown that are far from the AI techniques used in the first video games, which were defined by rule-based systems, that constitute the most basic algorithms in AI. More complex and innovative structures such as Deep learning or genetic algorithms are currently integrated. However, there is an AI technique called fuzzy logic that, despite not being as popular as neural networks, is very useful for this type of development and will be described below.

\subsection{Fuzzy Logic}
Fuzzy Logic is advisable for solving many problems where the uncertainty is a main actor, for instance the uncertain multi-criteria decision making problem \cite{chen2020choquet}. It is broadly used in controllers where it is hard to obtain a mathematical model, or systems that need to work with some vagueness or ambiguousness. Some applications created with fuzzy logic would be medical applications \cite{nimri2014artificial}, daily life applications as the washing machine \cite{gupta2017applications}, natural interaction with face recognition \cite{basak2018face} and posture recognition \cite{brulin2012posture} and lastly smart cities. Perhaps the last one could be disconcerting, notwithstanding the growth of cities has generated various problems in them. The administration of aspects such as waste collection, environmental pollution, traffic, energy distribution or water, becomes an increasingly complex task. Fuzzy logic has contributed to energy efficiency in smart cities \cite{krishna2018fuzzy}.
 
\subsubsection{Fuzzy Logic in Games}
Fuzzy logic, like other AI techniques cited previously, has also contributed to video games. In special needs, games have been developed specially for autistic people. In \cite{khabbaz2019adaptive} FL is used in order to rate the social skills in autistic children and adapt the level of the game to them. In the diagnoses of a patient's autism level, the patient has to carry out activities that will determine that level. The authors of this work \cite{iyer2017assess} have designed a game that is able to calculate the player's autism level using FL. Nevertheless, apart from carrying out work with people with autism, serious games have also been designed for people with physical impairments, such as the example of this game called ReHabGame \cite{esfahlani2017adaptive} where a series of activities are conducted for the upper and lower limbs, in order to evaluate and improve motor and sensory faculties in patients who have neuromuscular problems. This serious game allows the user to interact with Kinect in order to obtain information on the movement of their body during the performance of tasks and thus the fuzzy logic system will be able to extract and analyze the data corresponding to the positions of the users to satisfy their needs. In addition to ReHabGame, other serious games with different themes have also been developed, such as this serious game, in which the authors have studied the variables of throttle position sensor, engine rotation speed and car speed to determine by FL whether the driving style is efficient and in this way improve driver behavior in front of the steering wheel \cite{massoud2018fuzzy}. The authors expanded the work done integrating FL and Random Forest in order to know which algorithm was best for this type of serious game. The results indicated that the advantage of FL was that it produced understandable linguistic feedback while Random Forest predicted fuel consumption more accurately \cite{massoud2019exploring}. Another work that is worth mentioning for its novelty is the study conducted in \cite{lara2018induction} which aims to create a game to induce emotions in students with the aim of improving their learning process through inductive control. This system integrates fuzzy logic to analyze the performance and emotional state of the players through a voice analysis.

\subsection{AI Plugins for Unity}
Recently the Unity engine has finally embraced artificial intelligence, adding Machine Learning (ML) to its pathfinding tool. The most outstanding open-source plugin that this game engine has it is called ML-Agents. This plugin enables games and simulators to become a training environment to train artificial intelligence agents through reinforcement learning \cite{karangiya2019training}. Anyhow, if a developer wants to implement another technique the first stop is the Unity's Assets Store, filled with thousand of plugins. But among all the variety of tools available, those intended for implementing AI are scarce, and only one of them is based on fuzzy logic. This plugin is just a Fuzzy Logic layer for another tool aimed to build fighting games. The best solution is using an external library called AForge.net (AForge.NET). Written in C\#, is an AI library that includes an API for working with Neural Networks, Artificial Vision, and Machine Learning, among others. Given the broad scope of the library, the Fuzzy Logic implementation is shallow but easy to use. It has no Unity integration, so it lacks of GUI and makes it harder to use by staff like designers. The project is open source with a LGPLv3 license, so it can be easily extended, but our tool already has most of the elements it lacks like more membership functions, operators, or defuzzification methods. Ithaca was built from scratch thinking in Unity, so it implements components that can be attached to game objects. It also has a GUI for defining a system and debugging it without writing a single line of code. All these makes our solution more complete, more powerful and flexible, and thanks to its integration with Unity, easier to use.

\section{Ithaca} \label{proposal}
\subsection{What is Ithaca}
Ithaca is a tool for integrating a Fuzzy Logic Inference System (FIS) on any project developed with the Unity engine. It is written in C\# but it is not compiled, so all code is available to the developer for any needed modification. This makes the tool platform-independent, allowing to use it in any device among the ones available in Unity. Its core consist in a Fuzzy Rule Based System (FRBS) that evaluates a set of rules which are formed by fuzzy sets and fuzzy logic. 

\subsection{Fuzzy Logic and Inference Engine}
The fuzzy inference system implemented in Ithaca uses the Mamdani model, where the consequent of the IF-THEN rules is a fuzzy statement like this:

\[\textnormal{IF X is A and Y is B THEN Z is C} \]

This way of writing rules is natural and intuitive, making it more suitable for developing systems where we are trying to simulate human decision making. This model is less efficient than the Sugeno model, where the consequent of the rules are algebraic expressions \cite{Kaur12} so the defuzzifying step is faster to compute. For defuzzifying Mamdani FIS we implemented the next methods:

\begin{itemize}
\item Centroid or Centre of Gravity (COG):
	\[ 
	COG = \frac{\int_{min}^{max} x\mu (x) dx }{ \int_{min}^{max} \mu (x) dx}
	\]
\item Centroid for Singleton (COGS):
  \[ 
  COGS = \frac{ \sum_{i} x_{i}\mu_{i} }{ \sum_{i} \mu_{i} }
  \]
\item Bisector or Centre of Area (COA):
  \[ 
  COA = u', \int_{min}^{u'} \mu(u)du =  \int_{u'}^{max'} \mu(u)du
  \]
\item Mean of Maximum (MOM):
  \[
  MOM = \frac{\int_{ A_{max} }^{} x dx}{\int_{ A_{max} }^{}dx}
  \]
\item Right Most (RM):
  \[ 
  RM = \lbrace x | x > x',  \forall x' \in A_{max} \rbrace
  \]
\item Left Most (LM):
  \[ 
  LM = \lbrace x | x < x',  \forall x' \in A_{max} \rbrace
  \]
\end{itemize}

For the fuzzifying interface we wrote twelve membership functions, being some of them specific cases from more general functions, but broadly used. Regarding the logical connectives for the rules' elements, conjunction can be defined by a set of different t-norm methods. The same happens to the union, which may be defined using one of various t-conorms. All the methods implemented are:

\begin{itemize}
\item Membership functions:
	\begin{itemize}
    \item Singleton.
    \item Piecewise.
    \item Triangle.
    \item Trapezoid.
    \item Grade.
    \item Reverse grade.
    \item Gaussian.
    \item Double Gaussian.
    \item Bell.
    \item Cosine.
    \item Sigmoidal.
    \item Difference of sigmoidals.
    \end{itemize}
\item T-Norm (AND operation):
    \begin{itemize}
    \item Min (G\"odel-Dummett).
    \item Product
    \item Bounded Difference (\L ukasiewicz).
    \end{itemize}
\item T-CoNorm (OR operation):
    \begin{itemize}
    \item Max.
    \item ASUM.
    \item BSUM.
    \item NSUM.
    \end{itemize}
\end{itemize}

\subsection{Architecture}
The system architecture is divided in three subsystem; the core, the graphics subsystem, and the parser. The core subsystem has three main components (Figure \ref{fig:architecture}): 

\begin{figure}[!htb]
\centering
\includegraphics[width=\columnwidth]{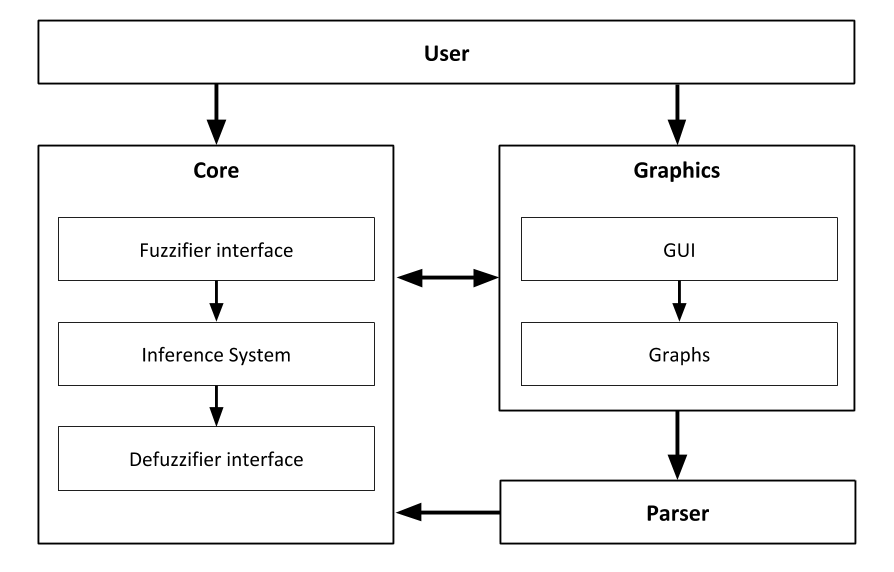}
\caption{System architecture}
\label{fig:architecture}
\end{figure}

\begin{enumerate}
\item The fuzzifier interface: this is where the crisp input values are turned into fuzzy values depending on the membership functions.
\item The inference system: all the rules are evaluated here. The antecedents determine the value of the consequences and the result is a fuzzy value.
\item The defuzzifier interface: the outputs are turned into crisp values in order to be used in the game.
\end{enumerate}

\begin{figure}[!h]
\centering
\captionsetup{justification=centering}
\includegraphics[width=\columnwidth]{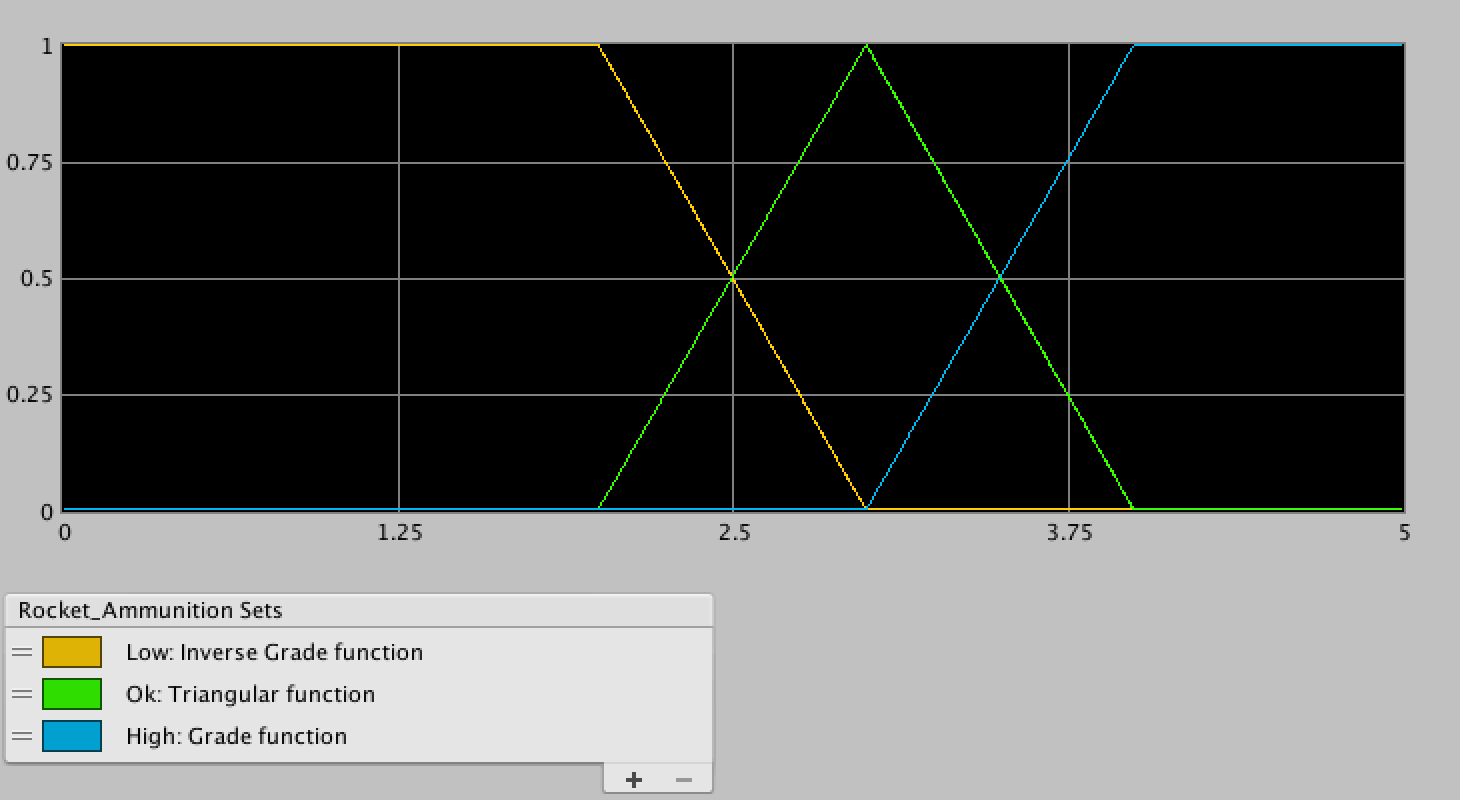}
\caption{Linguistic variable editor}
\label{fig:lveditor}
\end{figure}

The graphics subsystem is formed by the classes for the GUI and an additional set of classes, in a separated namespace, for drawing the graphs for the membership functions (Figure \ref{fig:lveditor}). Unity has no built-in system for drawing graphs, so we had to build our own plugin. We made it decoupled from the rest of the GUI so it could be used in other cases. At last, the parser subsystem translates the FCL files to the Ithaca inner format.

The inference engine structure is based in the FCL's definition, where a system is represented by a Function Block (FB), which in turn is defined by a set of Linguistic Variables (LV) as inputs and outputs, and a set of rules or Rule Block (RB). Each LV represents a domain of knowledge, and if declared as input has as many fuzzy sets as needed. Each fuzzy set has a membership function that defines the degree of correspondence of an input value to this set.  On the other side, if a LV is declared as output, it has an additional defuzzifier method to translate the fuzzy values to crisp values. The rules set is changeable in runtime, making it possible to adapt the knowledge base depending of the needs. 

Once a system is defined using the API, GUI, or a FCL file, the inference engine can be called during gameplay asking for an output depending on the current value, or with a new value.

\begin{figure}[htb]
\centering
\includegraphics[width=\columnwidth]{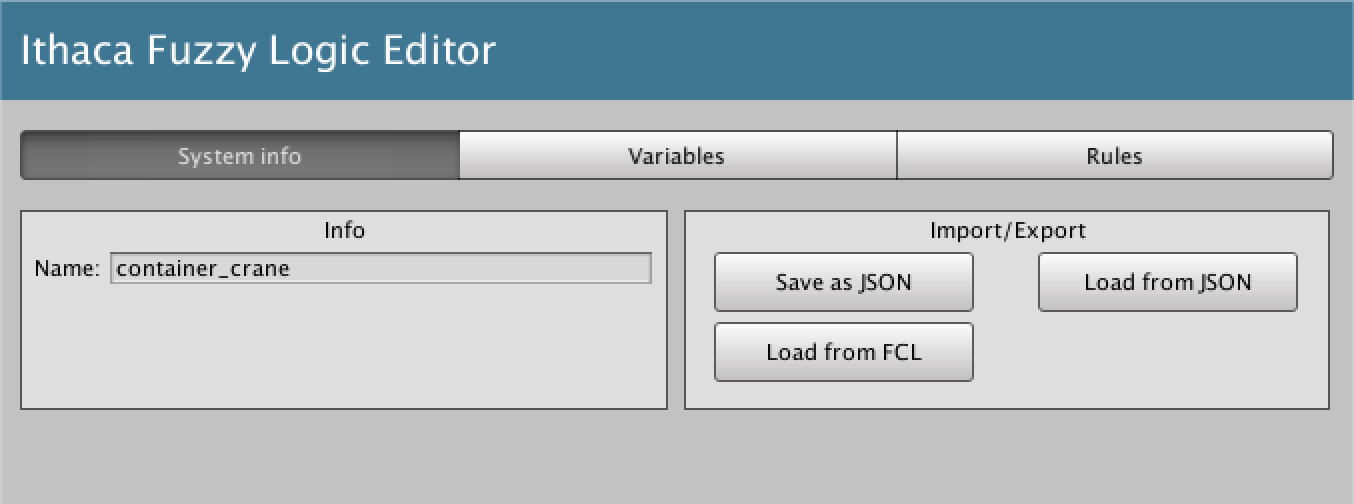}
\caption{Ithaca window}
\label{fig:window}
\end{figure}

The GUI has a main window with three tabs (see Figure \ref{fig:window}): 
\begin{itemize}
    \item One for creating/loading a system
    \item Another one for defining input and output variables
    \item The last one to write sets of rules
\end{itemize}    

The first tab lets the user create a new system or import one already defined. The systems can be exported to JSON format, and be loaded back from JSON and also from an FCL file. This lets the users save and exchange systems in a format trackable by version control software. Once the system is created, an asset file is created in the Unity project. This is a binary file with a serialization of the full system. If the asset file is selected, it can be edited from this window. The second tab is for declaring the input and output linguistic variables. Once the LV is created, it can be edited in order to add the fuzzy sets and its membership functions (Figure \ref{fig:lveditor}). The user can declare the limits of the LV and its default value.

\begin{figure}[!htb]
\centering
\includegraphics[width=\columnwidth]{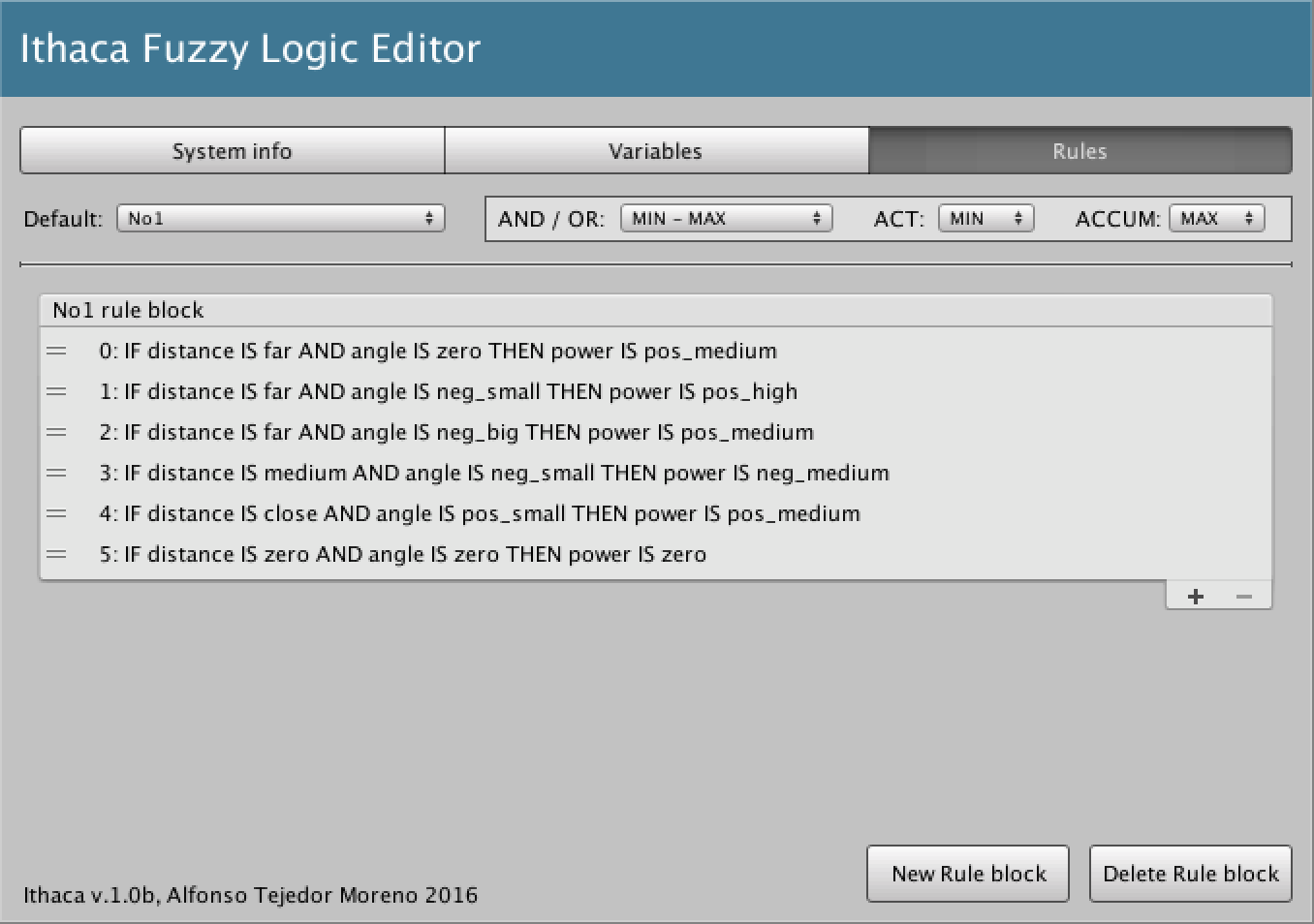}
\caption{Rules editor}
\label{fig:rules}
\end{figure}

Finally, in the third tab, the user can define sets of rules written in natural language (Figure \ref{fig:rules}). For each set of rules the user can decide the methods for the operations. This operations are:

\begin{itemize}
\item The methods for the logic operations AND/OR, also called aggregation operation. In order to satisfy De Morgan's law the methods must go in the following pairs:
  \begin{itemize}
  \item Min-Max
  \item Prod-ASUM
  \item BDIF-BSUM
  \end{itemize}
\item The methods for the activation operation. This operation translates the result of the antecedent to the consequent.
	\begin{itemize}
  	\item Min
    \item Prod
  	\end{itemize}
\item The method for the accumulation operation. This operation takes the results from the consequent of all fired rules and combines its values to obtain one output.
    \begin{itemize}
    \item Max
    \item BSUM
    \item NSUM
    \end{itemize}
\end{itemize}

If the used defines more than one set of rules, a default one must be choosen. As we said, the RB can be changed during runtime.

\begin{figure}[htb]
\centering
\includegraphics[height=9cm]{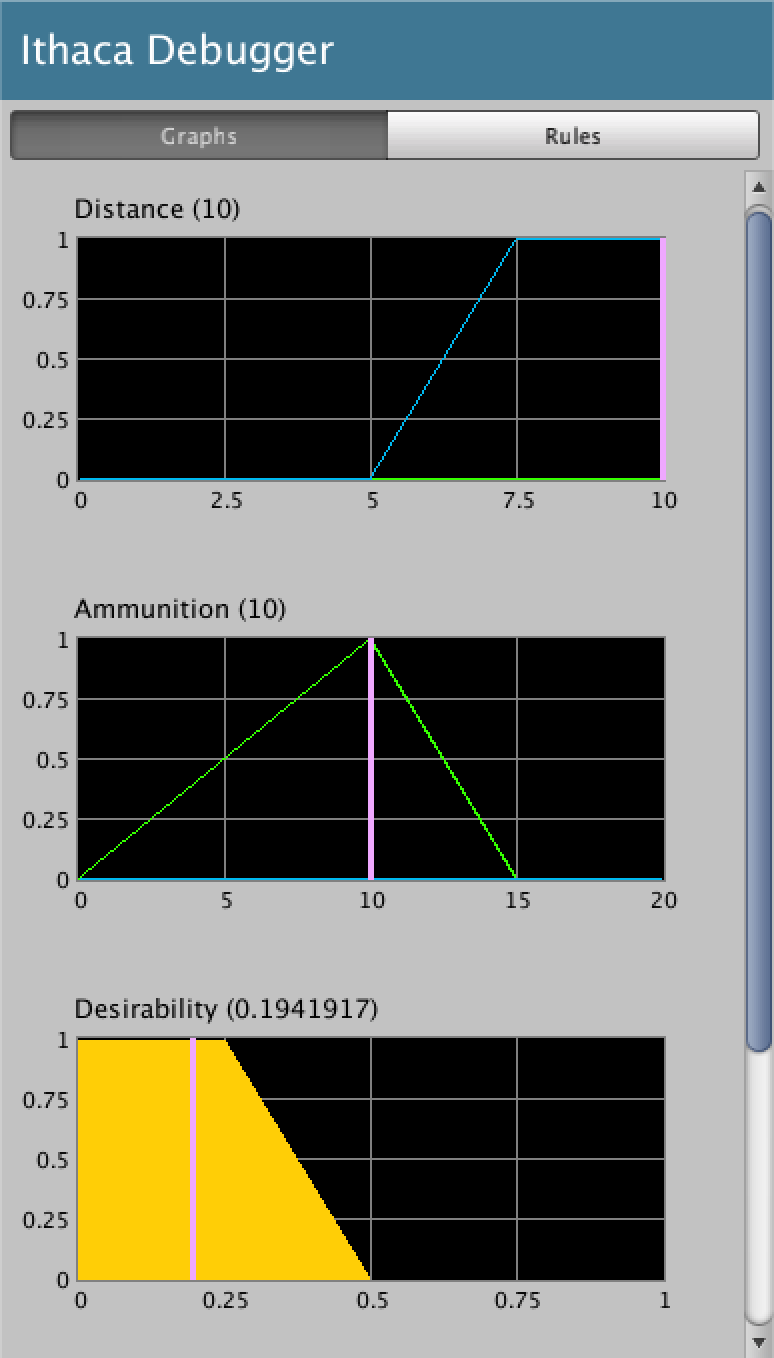}
\caption{Debug window}
\label{fig:debug}
\end{figure}

There is also an additional window, the debugger. It lets the developer debug both the graphs values and the rules that fired and its values. (Figure \ref{fig:debug}).

\subsection{Features}
The main features of the Ithaca tool are:

\begin{itemize}
\item A powerful GUI that allows to define a full system using graphs.
\item The inference system used is Mamdani, which uses fuzzy sets as outputs for the rules. This way the output is easier to use by the developer in most cases.
\item Covers the most used membership functions, defuzzifier methods, and operation expressions. The user can define new functions, methods, or expressions.
\item It is not compiled, thus the user can modify the code as needed within Unity.
\item A graphical debugging system that allow to check in real time the input and output values, as well as the values of every rule and if it was fired. 
\item It can be used to build games in any platform available in Unity. 
\end{itemize}

\section{Results} \label{results}
\subsection{Crane}
\begin{figure}[htb]
\centering
\includegraphics[width=\columnwidth]{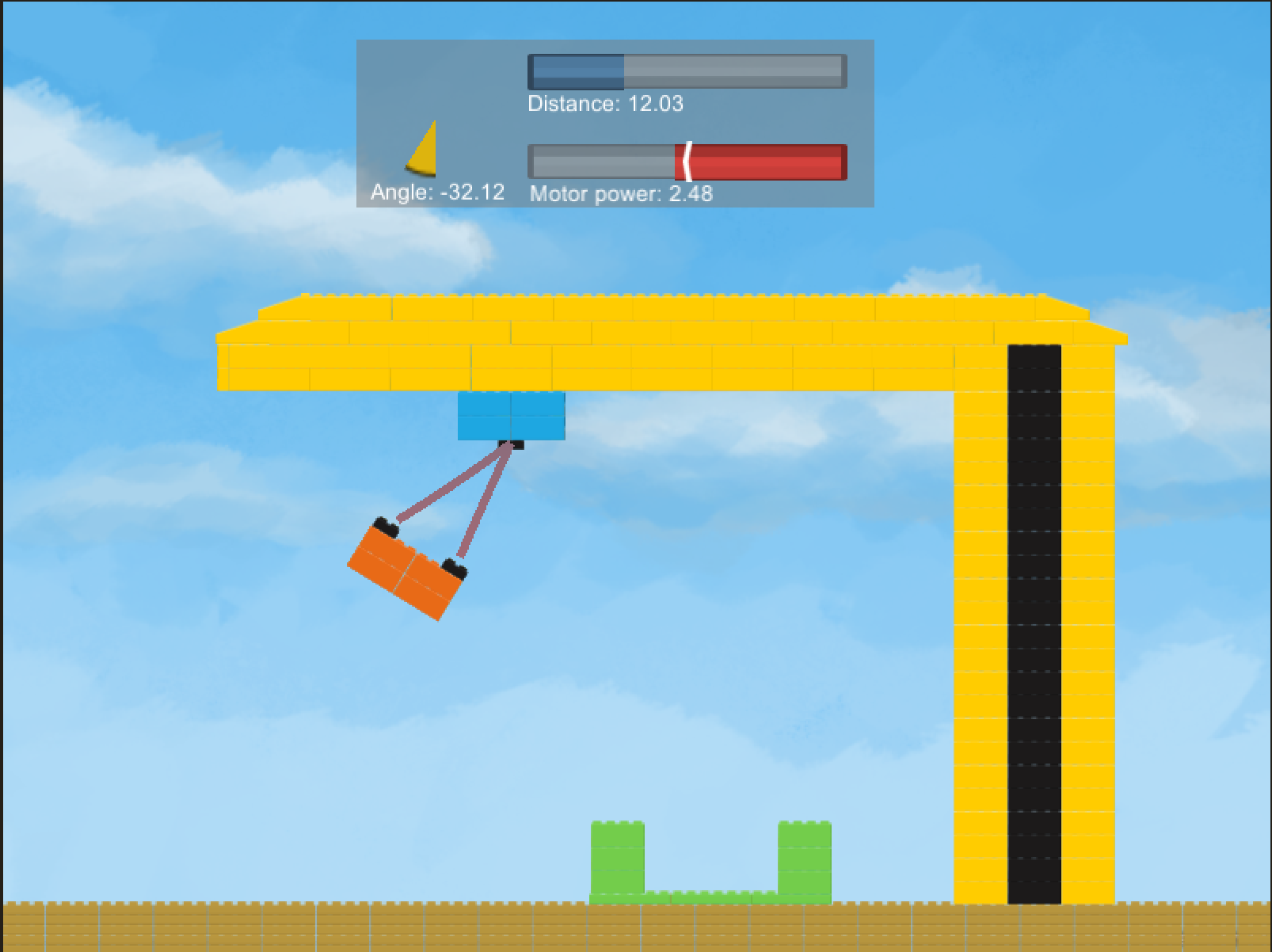}
\caption{Crane simulation}
\label{fig:crane}
\end{figure}

The goal of this test was to build a complex system using our tool, and to check the compliance with the FCL standard. So we decided to use the crane system defined in the FCL standard \cite{iec}, that controls container crane that loads and unloads containers from ships. The controller must moves the containers taking care of the speed and angle(Figure \ref{fig:crane}). The rules of the system were:

\begin{enumerate}

\item IF distance IS far AND angle IS zero THEN power IS pos\_medium.
\item IF distance IS far AND angle IS neg\_small THEN power IS pos\_high.
\item IF distance IS far AND angle IS neg\_big THEN power IS pos\_medium.
\item  IF distance IS medium AND angle IS neg\_small THEN power IS neg\_medium.
\item IF distance IS close AND angle IS pos\_small THEN power IS pos\_medium.
\item IF distance IS zero AND angle IS zero THEN power IS zero.
\end{enumerate}

The test demonstrated both the compliance with the FCL standard and the usefulness of the tool for building complex system using natural knowledge provided by specialists like a crane operator.

\subsection{Race car}
The objective in this test was to develop an adversary AI that looks natural and realistic, and to develop it entirely with the GUI. Many times it is important to keep the illusion that we are playing against a human to engage with the player, even if it makes our AI less optimal. The vagueness of the Fuzzy Logic makes it ideal for developing systems that act as imprecise as a human would. In this case we built two race cars, both with the same sensors and almost the same set of rules.  But one, called "Classic car", used classic logic, while the "Fuzzy car" used our inference system. The knowledge base used was:

\begin{enumerate}
\item IF Front IS Normal THEN Steering IS Straight,
  Speed IS Somewhat Fast.
\item IF Front IS Far THEN Steering IS Straight, Speed IS Very Fast.
\item IF Left IS Close THEN Steering IS Right.
\item IF Right IS Close THEN Steering IS Left.
\item IF VeryLeft IS Close THEN Steering IS VeryRight.
\item IF VeryRight IS Close THEN Steering IS VeryLeft.
\end{enumerate}

\begin{figure}[htb]
\centering
\includegraphics[width=\columnwidth]{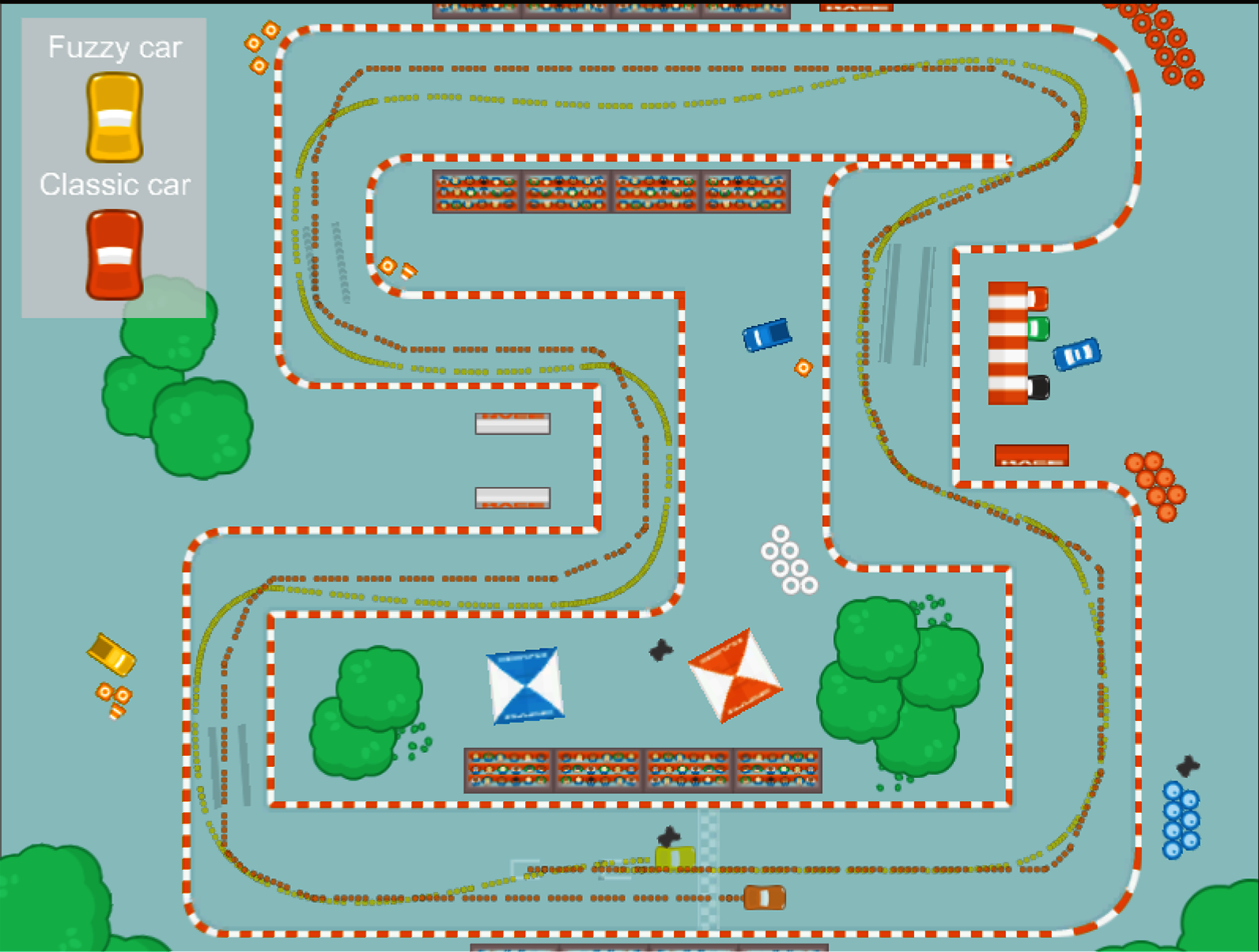}
\caption{Paths followed by AI cars}
\label{fig:carsPaths}
\end{figure}

The results were that the fuzzy car was faster and its trajectory was much more natural (Figure \ref{fig:carsPaths}). As for the use of the GUI, the system revealed itself as faster than using the API.

\section{Conclusions} \label{conclusions}

After the experience of developing those tests and some more small games, the power of our tool was clear. Its installation and integration on the project are really simple and straightforward, making the first step as easy as possible for any developer, especially the ones less experienced. This point is noteworthy as many of the users of Unity are not seasoned programmers.

The use of the GUI or the FCL makes building a Fuzzy Logic AI more approachable for small teams or solo developers, who can spend more time designing the behaviour of the agents rather than writing code. The use of natural language for writing the rules and defining the variable makes it easier to involve staff that does not know how to code but has a great knowledge of the matter. This allows these teams to focus on what is perceived by the players, the gameplay and the way of acting of its artificial intelligence.

Not only is it easy to develop, but it is also very simple to debug and maintain. Using natural language the developer is able to understand what is happening in the system at any moment and know what to expect from it. The debugging window is additional help on this task, as the user can read the status of the system and its expected outcomes in a visual way.

\subsection{Future Work}
\begin{itemize}
\item To implement Combs method for reducing the number of rules needed by combining them.
\item To add the option to create Fuzzy Finite State Machines (FuFSM). The FSM is a very popular technique among game developers and the use of FL can provide a degree of uncertainty very valuable in some cases.
\item To generate the knowledge base implementing Adaptive Network-based Fuzzy Inference System (ANFIS)\cite{anfis}. This would allow creating more powerful systems thanks to the use of NN.
\item To implement Takagi-Sugeno method for inference.
\item To implement the new Fuzzy Markup Language (FML), based on XML \cite{ieee1855ieee}.
\end{itemize}

\section*{Acknowledgment} \noindent This work was funded by the EU ERDF and the Spanish Ministry of Economy and Competitiveness (MINECO) under the Project TIN2017-83964-R. This work also received funding from the CEiA3 and CEIMAR consortium. 

\bibliographystyle{IEEEtran}
\bibliography{references.bib}

\end{document}